\documentclass[a4paper]{article}
\pdfoutput=1

\usepackage[english]{babel}
\usepackage[utf8]{inputenc}
\usepackage[T1]{fontenc}

\usepackage{amsfonts,amssymb,amsmath,amsthm}
\usepackage{graphicx}
\usepackage{caption}
\usepackage[position=b,justification=Centering,singlelinecheck=false]{subfig}
\usepackage[usenames,dvipsnames]{xcolor}
\usepackage{pgfplots}
\pgfplotsset{compat=newest}
\usepackage{pgfplotstable}
\usepgfplotslibrary{fillbetween}
\usepackage{tikz}	
\usepackage{empheq}
\usepackage[preprint]{spconf-arxiv}
\graphicspath{{pics/}}		
\toappear{To appear in {\it Proc.\ iTWIST2018,
                 November 20 - 22, 2018, Marseille, France}}

\newcommand{\bs}{\boldsymbol}

\newcommand{\bb}{\mathbb}
\newcommand{\cl}{\mathcal}
\newcommand{\ie}{\emph{i.e.}, }
\newcommand{\eg}{\emph{e.g.}, }
\DeclareMathOperator*{\argmin}{\arg\min}

\newlength{\sizeFig}
\newlength{\hspaceFig}
\newlength{\vspaceSec}
\newlength{\vspaceSecD}
\newlength{\vspaceEq}
\title{Compressive Sampling Approach \\ for Image Acquisition with Lensless Endoscope}
\name{S. Gu\'erit$^*$, S. Sivankutty$^\ddagger$, C. Scott\'e$^\ddagger$, J.~A.~Lee$^{*,\dagger,\mathsection}$, H. Rigneault$^\ddagger$, and L. Jacques$^{*,\mathsection}$%
	\thanks{$^\mathsection$ J.~A.~L. and L.~J. are Research Associates with the Belgian F.R.S.-FNRS. This study has been partly funded by \textsc{AlterSense}-MIS.}
}
\address{\small
    \tabular{c}
		$^*$ ICTEAM\\ 
		UCLouvain\\ 
		Louvain-la-Neuve, Belgium
	\endtabular
	\hskip 0.5in
	\tabular{c}
		$^\ddagger$ Institut Fresnel\\
		Aix-Marseille Université\\
		Marseille, France
	\endtabular
	\hskip 0.5in
    \tabular{c}
		$^\dagger$ MIRO\\ 
		UCLouvain\\ 
		Brussels, Belgium 
	\endtabular
}
\begin{document}

\maketitle
\begin{abstract}
The lensless endoscope is a promising device designed to image tissues \emph{in vivo}  at the cellular scale. The traditional acquisition setup consists in raster scanning during which the focused light beam from the optical fiber illuminates sequentially each pixel of the field of view (FOV). The calibration step to focus the beam and the sampling scheme both take time. In this preliminary work, we propose a scanning method based on compressive sampling theory. The method does not rely on a focused beam but rather on the random illumination patterns generated by the single-mode fibers. Experiments are performed on synthetic data for different compression rates (from 10 to 100\% of the FOV).
\end{abstract}
\section{Introduction}
\label{sec:introduction}
Nowadays, \emph{in vivo} imaging of tissues is performed everyday in hospitals (\eg magnetic resonance imaging or computed tomography). The spatial resolution of such devices is far from the cell resolution whose diameter typically ranges from 10 to 100 $\mu$m \cite{Milo2009}. Cell imaging is almost exclusively done \emph{ex vivo}. However, due to recent developments in microscopy and optics fields, \emph{in vivo} imaging of cells in their natural environment raises growing interest \cite{Andresen2016}. But in their review, Andresen \emph{et al.} \cite{Andresen2016} point out that the current devices suffer from physical limitations restricting the imaging depth to 1 mm. 
To overcome this issue, researchers aim to design miniaturized imaging systems with no opto-mechanical components at the fiber distal end, called \emph{lensless endoscopes}. At Institut Fresnel, researchers develop lensless endoscopes based on multicore optical fibers \cite{Andresen2013,Sivankutty2018}. Their ultrathinness (300 $\mu$m) makes the probes not much invasive and well suited to explore and image cells, \eg neurons, at depths beyond reach of other imaging systems.

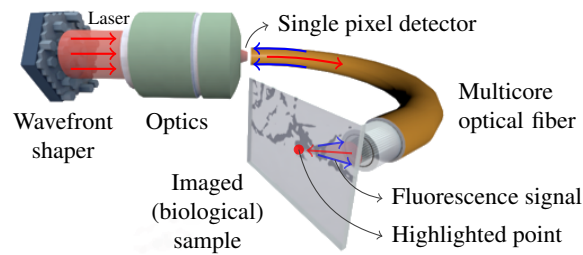
\begin{figure}
	\centering
	\begin{tikzpicture}[font=\small]
		\node at (0,0) {\includegraphics[width=0.39\linewidth]{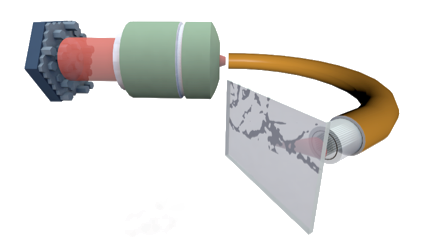}};
		\node [text width = 2 cm] at (4,0.2) {Multicore \\ optical fiber};
		\node at (-0.7,0) {Optics};
		\node [font=\scriptsize] at (-1.6,1.4) {Laser};
		\draw [->,gray!50!black] (0.2,1) to [bend left] (0.5,1.3) node [anchor=west,black] {Single pixel detector};
		\node [text width= 1.5cm, align = center] at (-0.3,-1.2) {Imaged \\ (biological) \\ sample};
		\node [text width= 2cm, align = center] at (-2.2,-0.2) {Wavefront \\ shaper};
		
		% Laser arrows
		\draw [red,thick,->] (-2.1, 1.12) to (-1.5,1.12);
		\draw [red,thick,->] (-2.1, 0.92) to (-1.5,0.92);
		\draw [red,thick,->] (-2.1, 0.72) to (-1.5,0.72);
		
		% Optical fiber arrows
		\draw [red,thick,->] (0.48, 0.88) to [out = 0, in = 170] (1.48,0.78);
		\draw [blue,thick,->] (1, 0.74) to [in = 0, out = 170] (0.3,0.78);
		\draw [blue,thick,->] (1, 0.95) to [in = 0, out = 170] (0.3,0.99);
		
		% Light on biological sample
		\draw [red,fill=red,opacity=0.8] (0.9,-0.35) circle (0.06) node (B) {};
		\draw [red,thick,->,opacity=0.6] (1.6,-0.4) to (1,-0.36);
		
		\draw [blue,thick,->,opacity=0.6] (1.15,-0.3) to (1.5,-0.25);
		\draw [blue,thick,->,opacity=0.6] (1.15,-0.42) to node [pos=0.5] (A) {} (1.6,-0.55);
		\draw [->,gray!50!black] (A.center) to [bend right] (2,-1) node [anchor = west,black] {Fluorescence signal};
		\draw [->,gray!50!black] (B.center) to [bend right] (2,-1.5) node [anchor = west,black] {Highlighted point};
		
	\end{tikzpicture}
	\caption{\label{fig:endoscope_principle}Lensless endoscope principle in raster scanning mode (\ie the light beam is shaped to focus on one point). Source: Institut Fresnel$^2$.}
\end{figure}

The principle of point-scanning or \emph{raster scan} (RS) with a lensless endoscope is illustrated in Fig.\,\ref{fig:endoscope_principle}. The entire device consists of a wavefront shaper, here a deformable mirror, an optics part to focus light into fibers, and a multicore fiber. The optical fiber inner part - dedicated to excitation - is made of single-core fibers displayed in a golden spiral shape \cite{Sivankutty2018}. The outer part of the fiber is made of air and consitutes a ``double-clad'' dedicated to signal collection. 
During the calibration step, specific sequences of proximal wavefronts are designed such that the light from all cores will focus on specific positions of the sample. During the acquisition, the light beam focuses sequentially on each (discretized) position of the field of view (FOV), generating a fluorescence signal. This signal travels back in the outer part of the optical fiber and is measured by an external sensor. The image is then reconstructed pixel by pixel. 

The design of single-core fibers spatial arrangement as well as the calibration step are essential to get a focused light beam, \ie a focused point spread function (PSF). In practice, (\emph{i}) the focus of the light at the distal end of the fiber is imperfect (with secondary modes in the illumination pattern) and (\emph{ii}) noise corrupts the observations. Post-processing step is necessary.

But what happens if we relax the constraint of focusing the light beam? In this case, the biological sample will be illuminated by random light patterns (\emph{speckles}). This acquisition process is then related to the principle of the single-pixel camera \cite{Duarte2008} and thus to the theory of compressive sampling (CS) \cite{Candes2008,Jacques2010a}. In compressive single-pixel imaging, the image is observed through its correlations with a grid of randomly oriented micro-mirrors. If the image structure can be represented in a sparse way, only a limited number of measurements, or correlations, are necessary to achieve good reconstruction quality. 

In this preliminary work, we investigate the compressive sampling approach for image acquisition with a lensless endoscope using random illumination patterns. For different compression rates (\ie the number of measurements \emph{versus} the number of pixels), we will compare its performance to the conventional raster scanning method. 

\section{Acquisition model}
Good knowledge of the forward model relating observations $\bs y$ to original image $\bs x$ is necessary to reconstruct $\bs x$ from $\bs y$.

\subsection[Acquisition of N observations]{Acquisition of $\bs N$ observations}Let $\bs x \in \mathbb{R}^N$ be the original vectorized $N$-pixels image resulting from the discretization of the sample located in the FOV. Let us assume that the PSF of the optical device (with or without the calibration step) is spatially invariant, \ie the pattern illuminating the sample remains identical when it is shifted to different locations of the FOV. In this case, observations $\bs y \in \mathbb{R}^N$ result from the linear convolution of $\bs x$ with $\bs h \in \mathbb{R}^N$, the discrete version of the PSF,%\vspace{-1mm}
\begin{equation}%\vspace{-1mm}
	\bs y = \bs h \otimes \bs x + \bs n, \quad n_i \sim_{\rm i.i.d.} \cl N(0,\sigma_n^2)
	\label{eq_basic_forward-model}
\end{equation}
where $\bs n \in \bb R^N$ is an additive noise. 
When there is no calibration, instead of illuminating all spatial positions of the sample with the same light pattern, we could introduce diversity in the observations by randomly alternating between $P$ speckles,%\vspace{-1mm}
\begin{equation}%\vspace{-1mm}
	\bs y = \sum_{i=1}^P \bs S_{\Omega_i}(\bs h_i \otimes \bs x) + \bs n,
	\label{eq_forward-model_random-patterns}
\end{equation}
where $\bs S_{\Omega_i} \in \bb R^{N\times N}$ is a masking operator, \ie $(\bs S_{\Omega_i}\bs u)_j = u_j$ if $j\in\Omega_i$ and 0 otherwise. The scanning of the whole FOV requires $\{\Omega_1,\dots,\Omega_P\}$ to be a partition of the set $\{1,\dots,N\}$.

\subsection[CS acquisition of M < N observations]{CS acquisition of $\bs{M < N}$ observations}As mentionned in the introduction, the CS acquisition model consists in acquiring fewer measurements (from random illumination patterns) compared to the actual size of the FOV. In this study, we acquire observations at the center of the FOV (see Fig.\,\ref{fig:results_images}) but random sampling is another possible choice. Eq.\,\eqref{eq_forward-model_random-patterns} becomes %\vspace{-1mm}
\begin{equation}%\vspace{-1mm}
	\bs y_\text{CS} = \bs R\left(\sum_{i=1}^P \bs S_{\Omega_i}(\bs h_i \otimes \bs x) + \bs n\right), \qquad \bs y_\text{CS} \in \bb R^M,
	\label{eq_forward-model_random-patterns_CS}
\end{equation}
where $\bs R \in \bb R^{M\times N}$ is a restriction operator such that $\forall \bs u \in \bb R^N$, $\text{supp}(\bs R \bs u)$ is a square of $M$ pixels located at the center of the FOV. For $P=1$ and $M=N$, the CS model reduces to \eqref{eq_basic_forward-model}.

\section{Image estimation}
In this work, we aim at finding the best estimate $\bs{\tilde x}$ for $\bs x$ from either full FOV acquisition or CS acquisition. In both cases, we can formulate this \emph{inverse problem} mathematically as a minimization problem based on the linear forward model given in~\eqref{eq_forward-model_random-patterns_CS}, and solving it using tools of convex optimization. But minimizing the mean square error of $\bs{\tilde x}$ is an ill-posed problem, particularly due to the presence of noise. To overcome this issue, a common way is to regularize the problem by adding some prior information about the signal of interest.

Estimate $\bs{\tilde x}$ is found by solving the following minimization,%\vspace{-1mm}
\begin{equation}%\vspace{-1mm}
	\bs{\tilde x} = \argmin_{\bs{\bar x} \in \bb R^N} \|\bs R \sum_{i=1}^P \bs S_{\Omega_i}(\bs h_i \otimes \bs {\bar x})-\bs y\|_2^2 + \rho\,\bs\Phi(\bs{\bar x}),
	\label{eq_inverse_problem}
\end{equation}
where $\bs\Phi(\bs{\bar x})$ is the regularization term and parameter $\rho > 0$ controls the trade-off between the two terms. The optimal value of $\rho$ depends on the noise statistics and is not known \emph{a priori}. In this work, $\bs\Phi$ is the total variation (TV) norm, defined as the $\ell_1$-norm of the image gradient magnitude \cite{Rudin1992}, and minimal for piecewise constant images. This regularization is widely used but can lead to staircasing effects on natural images \cite{Chambolle2010}. Priors like sparsity in the wavelet domain \cite{Kamilov2012,Almeida2013,Gonzalez2016a}, sparsity of the image edges \cite{Almeida2010} or including higher derivatives of the images \cite{Bredies2010} are more adapted and will be investigated in future work.

The algorithm used to solve \eqref{eq_inverse_problem} is the alternating direction method of multipliers \cite{Parikh2013}. It belongs to the family of proximal algorithms. Such algorithms are able to solve convex problems with non-smooth and non-differentiable objective function, like the $\ell_1$-norm. The optimal value of $\rho$ is estimated iteratively according to the rule described in \cite{Gonzalez2016a} and based on whiteness of the residual $(\bs R \sum_{i=1}^P \bs S_{\Omega_i}(\bs h_i \otimes \bs {\bar x})-\bs y)$ \cite{Almeida2013a}.

% Figure with results on real phantom
\begin{figure}
	\centering
		\subfloat[\label{fig:xgt_psf_a}]{\includegraphics[width=\sizeFig]{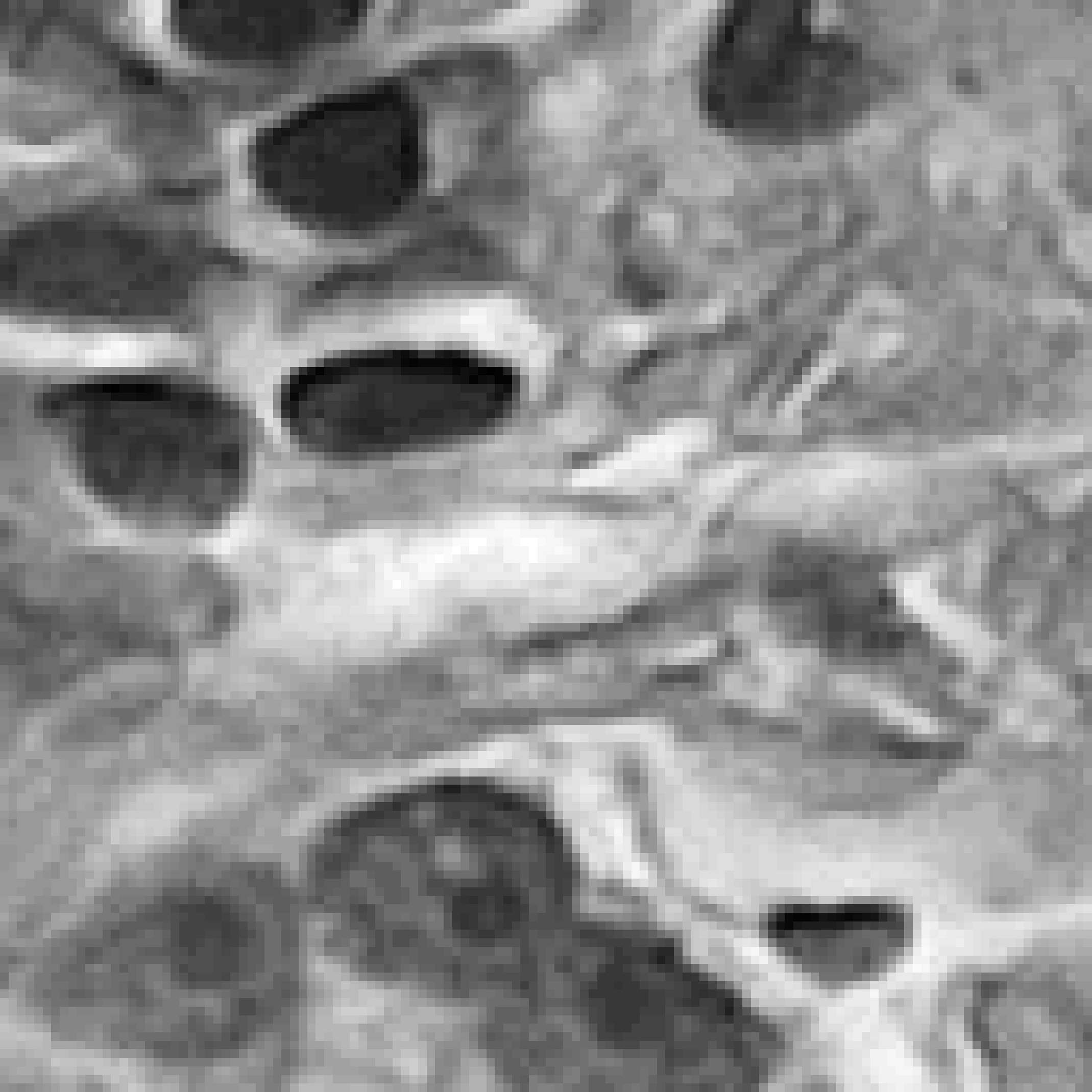}} \quad
		\subfloat[\label{fig:xgt_psf_b}]{\includegraphics[width=\sizeFig]{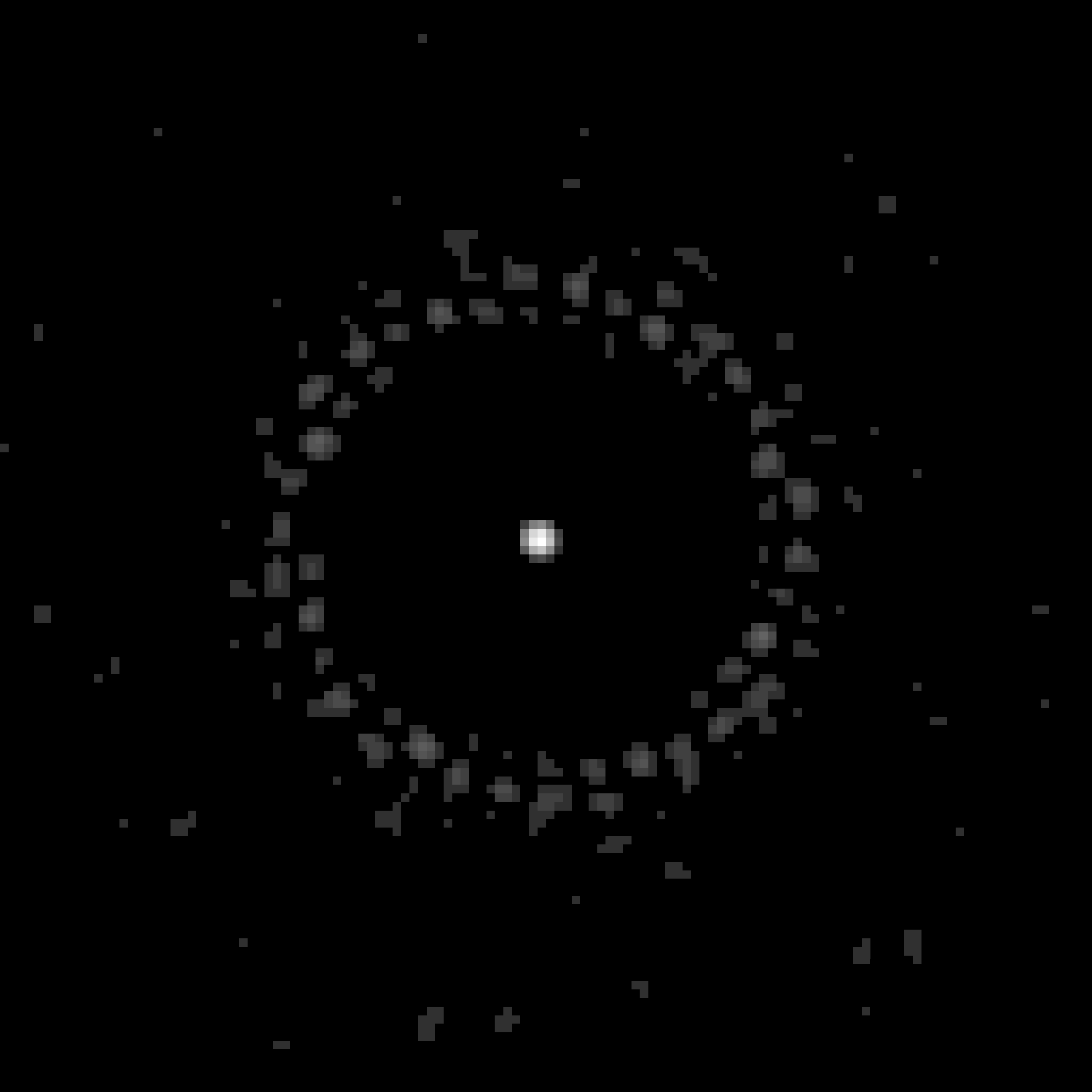}} \quad
		\subfloat[\label{fig:xgt_psf_c}]{\includegraphics[width=\sizeFig]{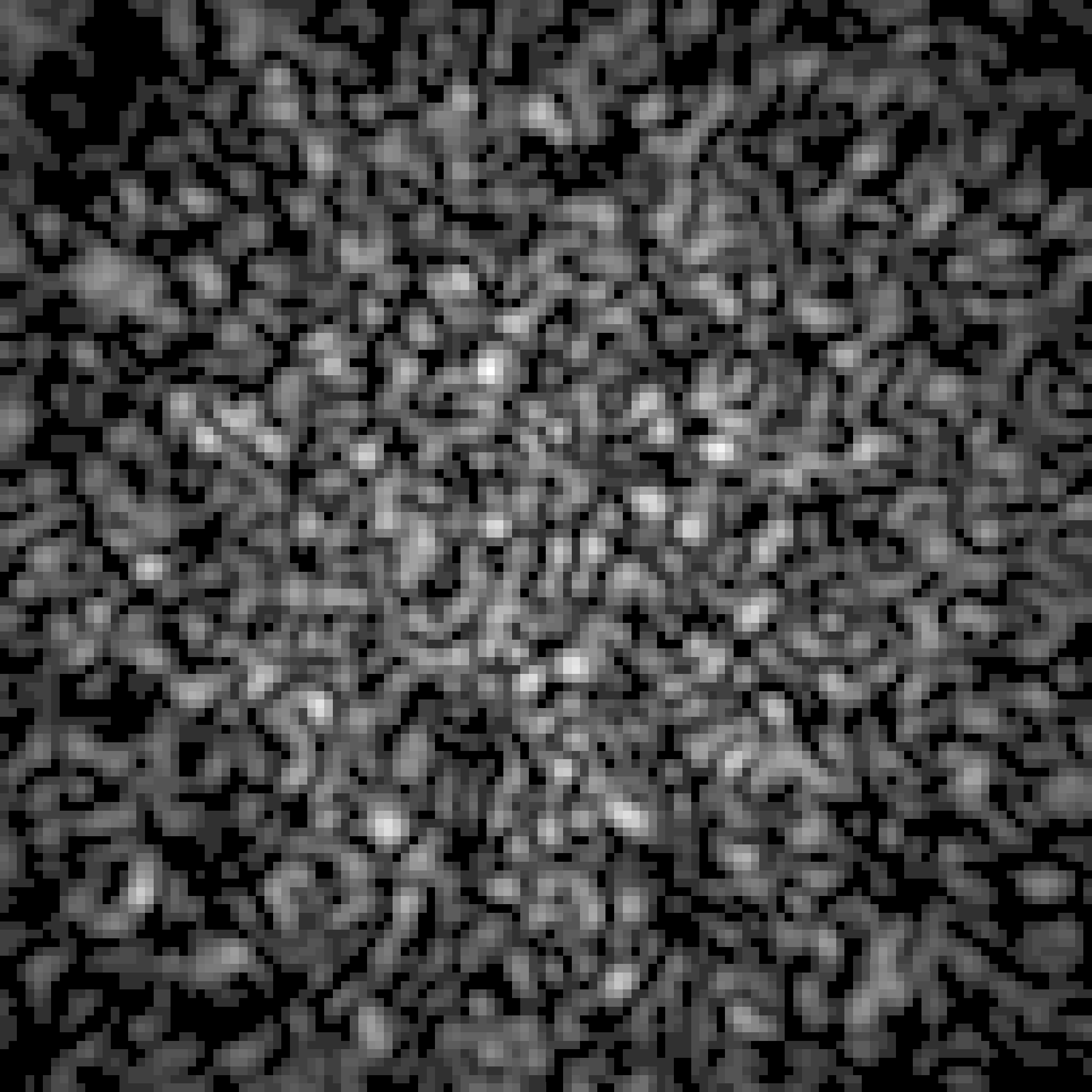}} 
	\caption{Original tissue image (left), simulated focused PSF (center), and one of the simulated random patterns (\emph{speckles}) obtained without fibers calibration (right). Center and right images were brightened for visualisation purpose.}
	\label{fig:xgt_psf}
\end{figure}

\section{Results and discussion}
\label{sec:results}

\subsection{Method}To perform our experiment, we used a $128\times 128$ tissue image from Matlab \cite{Matlab} (see Fig.\,\ref{fig:xgt_psf_a}). The focused PSF and the speckles are simulated with Matlab (see Fig.\,\ref{fig:xgt_psf_b} and \ref{fig:xgt_psf_c}) and then corrected for vignetting artifacts. Observations $\bs y$ are generated according to \eqref{eq_forward-model_random-patterns_CS} with a noise variance $\sigma^2_n$ such that the blurred signal-to-noise ratio (BSNR) is equal to 40 dB. The BSNR is defined as the SNR of the observations without noise corruption. The SNR between original image $\bs u \in \bb R^N$ and image $\tilde{\bs u} \in \bb R^N$ is $\text{SNR}(\tilde{\bs u},\bs u) \coloneqq 20\log_{10}{(\|\tilde{\bs u}\|_2/\|\bs u - \tilde{\bs  u}\|_2)}$.

\subsection{Experiment}The experiment consists in reconstructing $\bs x$ from $\bs y$ with conventional RS and CS approaches for different $M/N$ ratios corresponding to $\{0.1,0.2,\dots,1\}$. In the CS framework, the number of speckles $P\in\{1,2,4\}$. The measurements are acquired at the center of the FOV. We run 20 trials for each parameters setting.

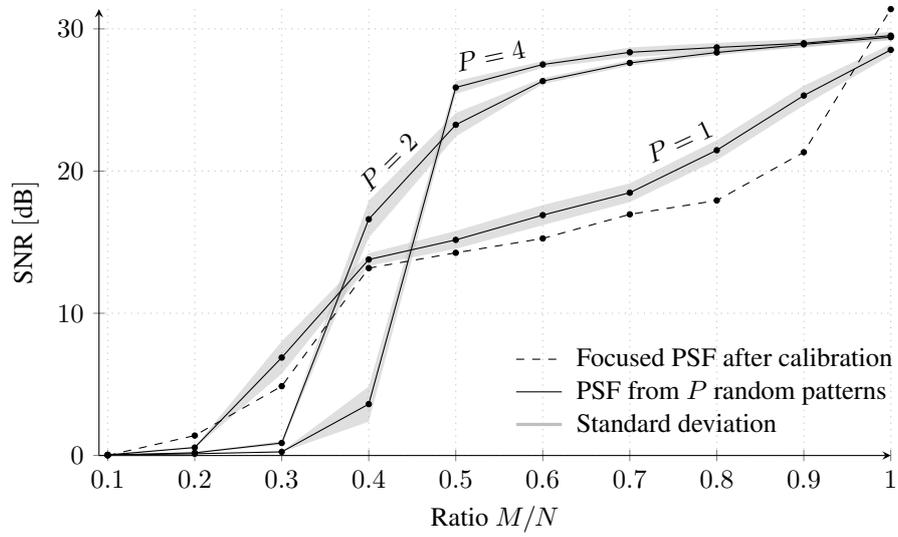
\begin{figure}
	\centering
	\begin{tikzpicture}
		\begin{axis}[
			grid = major,
			grid style = {dotted},
			height = 7.5cm,
			width = 0.75\linewidth,
			axis lines = left,
			xmin = 0.09,
			ylabel = {SNR $\left[\text{dB}\right]$},
	    	xlabel = {Ratio $M/N$},
			legend cell align = {left},
    		legend style = {at = {(1.02,0.02)},
      		anchor = south east,column sep = 0.1cm, draw = none, fill = none},
      		cycle list name=black white,
      		legend entries={Focused PSF after calibration,PSF from $P$ random patterns,Standard deviation}
			]
			
			\addlegendimage{no marks,black,dashed}
			\addlegendimage{no marks,black,solid}	
			\addlegendimage{no marks,gray!50,very thick,solid}	

			% Data importation	
			% ----------------							 	 
			\pgfplotstableread{data/NBDx_Tissue_CS_centered-samples_focusedKernel-M1_GaussianNoise-BSNR40_xPrior-TV_ADMM_20Trials.dat}\dataFocusMOne
			\pgfplotstableread{data/NBDx_Tissue_CS_centered-samples_speckleKernel-M1_GaussianNoise-BSNR40_xPrior-TV_ADMM_20Trials.dat}\dataSpeckleMOne
			\pgfplotstableread{data/NBDx_Tissue_CS_centered-samples_speckleKernel-M2_GaussianNoise-BSNR40_xPrior-TV_ADMM_20Trials.dat}\dataSpeckleMTwo
			\pgfplotstableread{data/NBDx_Tissue_CS_centered-samples_speckleKernel-M4_GaussianNoise-BSNR40_xPrior-TV_ADMM_20Trials.dat}\dataSpeckleMFour
			
			% Mean
			\addplot+ [dashed, mark=*, mark size=1pt, mark options={color=black,solid}] table[x expr = \thisrowno{0}, y expr = \thisrowno{1}, y index = 0] {\dataFocusMOne};

			% Random pattern (M=1)
			% --------------------
			% Standard deviation
			\addplot+ [name path = A, no marks, forget plot,draw=none,gray] table[x expr = \thisrowno{0}, y expr = \thisrowno{1}+\thisrowno{2}, y index = 0] {\dataSpeckleMOne};
			\addplot+ [name path = B, no marks, forget plot,draw=none,gray] table[x expr = \thisrowno{0}, y expr = \thisrowno{1}-\thisrowno{2}, y index = 0] {\dataSpeckleMOne};
			\addplot+ [opacity = 0.5, forget plot,fill = gray!50] fill between [of = A and B];
			
			% Mean
			\addplot+ [black, solid, mark=*, mark size=1pt, mark options={color=black,solid}] table[x expr = \thisrowno{0}, y expr = \thisrowno{1}, y index = 0] {\dataSpeckleMOne} node[rotate=25, anchor=south,pos=0.72,yshift=0.5ex] {$P=1$};

			% Random pattern (M=2)
			% --------------------
			% Standard deviation
			\addplot+ [name path = A, no marks, forget plot,draw=none,gray] table[x expr = \thisrowno{0}, y expr = \thisrowno{1}+\thisrowno{2}, y index = 0] {\dataSpeckleMTwo};
			\addplot+ [name path = B, no marks, forget plot,draw=none,gray] table[x expr = \thisrowno{0}, y expr = \thisrowno{1}-\thisrowno{2}, y index = 0] {\dataSpeckleMTwo};
			\addplot+ [opacity = 0.5, forget plot,fill = gray!50] fill between [of = A and B];
			
			% Mean
			\addplot+ [black,solid, mark=*, mark size=1pt, mark options={color=black,solid}] table[x expr = \thisrowno{0}, y expr = \thisrowno{1}, y index = 0] {\dataSpeckleMTwo} node[rotate=45, anchor=south,pos=0.66,yshift=0.5ex] {$P=2$};

			% Random pattern (M=4)
			% --------------------
			% Standard deviation
			\addplot+ [name path = A, no marks, forget plot,draw=none,gray] table[x expr = \thisrowno{0}, y expr = \thisrowno{1}+\thisrowno{2}, y index = 0] {\dataSpeckleMFour};
			\addplot+ [name path = B, no marks, forget plot,draw=none,gray] table[x expr = \thisrowno{0}, y expr = \thisrowno{1}-\thisrowno{2}, y index = 0] {\dataSpeckleMFour};
			\addplot+ [opacity = 0.5, forget plot,fill = gray!50] fill between [of = A and B];
			
			% Mean
			\addplot+ [black,solid,mark = *, mark size=1pt, mark options={color=black,solid}] table[x expr = \thisrowno{0}, y expr = \thisrowno{1}, y index = 0] {\dataSpeckleMFour} node[rotate=12, anchor=south,pos=0.9,yshift=0.3ex] {$P=4$};

		\end{axis} 
	\end{tikzpicture}
	\caption{Mean SNR (over 20 trials) of restored $128\times 128$ tissue image \emph{versus} $M/N$, \emph{i.e.}, the ratio between the number of observations and the number of pixels in the FOV. The BSNR of observations is 40\,dB.}
	\label{fig:SNR-vs-compression_centered-samples}
\end{figure}

\subsection{Results}
\begin{figure}
	\centering
		\subfloat{\raisebox{0.4\sizeFig}{\rotatebox[origin=c]{90}{\vphantom{$\bs{\tilde x}$}Observations $\bs y$}}} \quad
		\subfloat{\frame{\includegraphics[height=0.92\sizeFig]{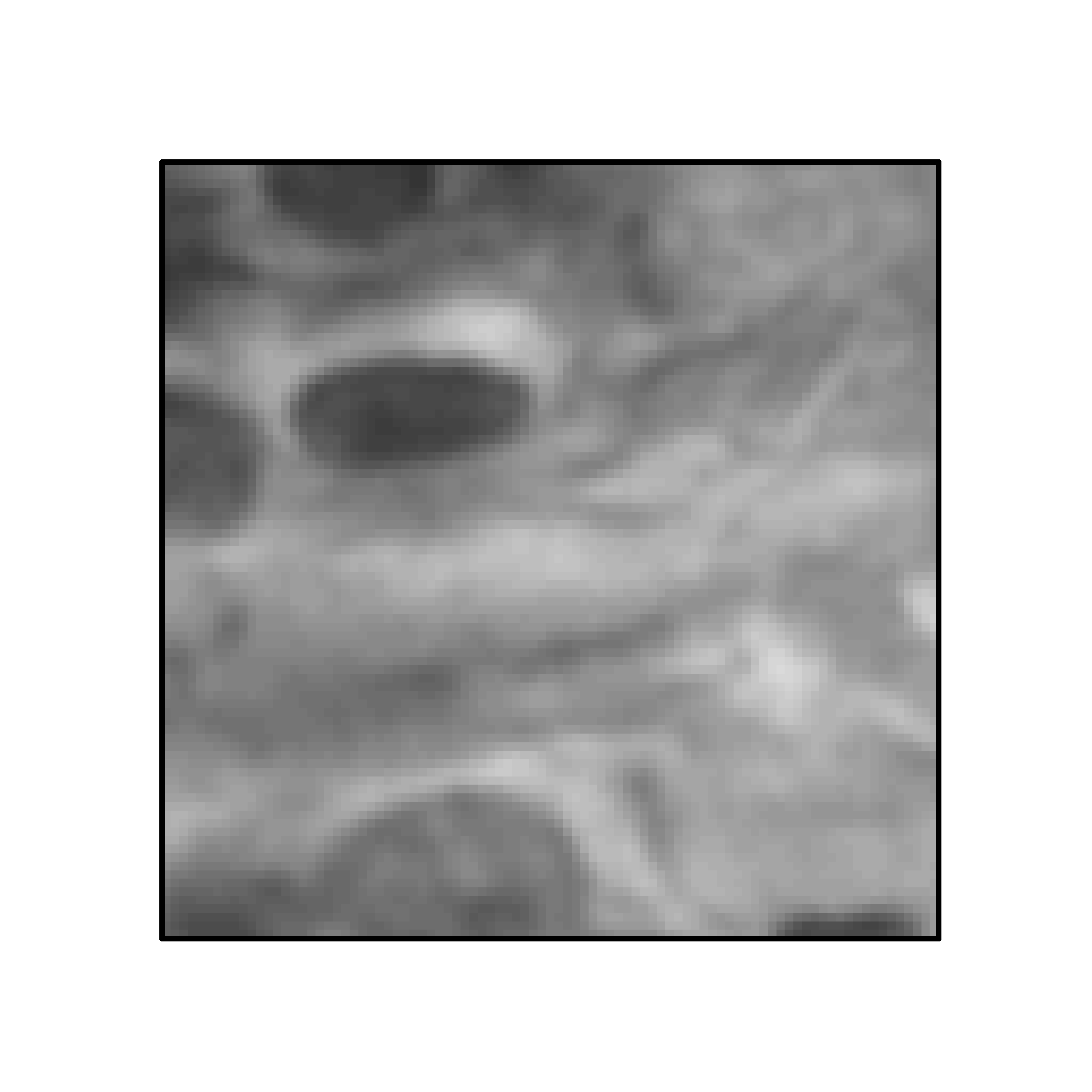}}} \quad
		\subfloat{\frame{\includegraphics[height=0.92\sizeFig]{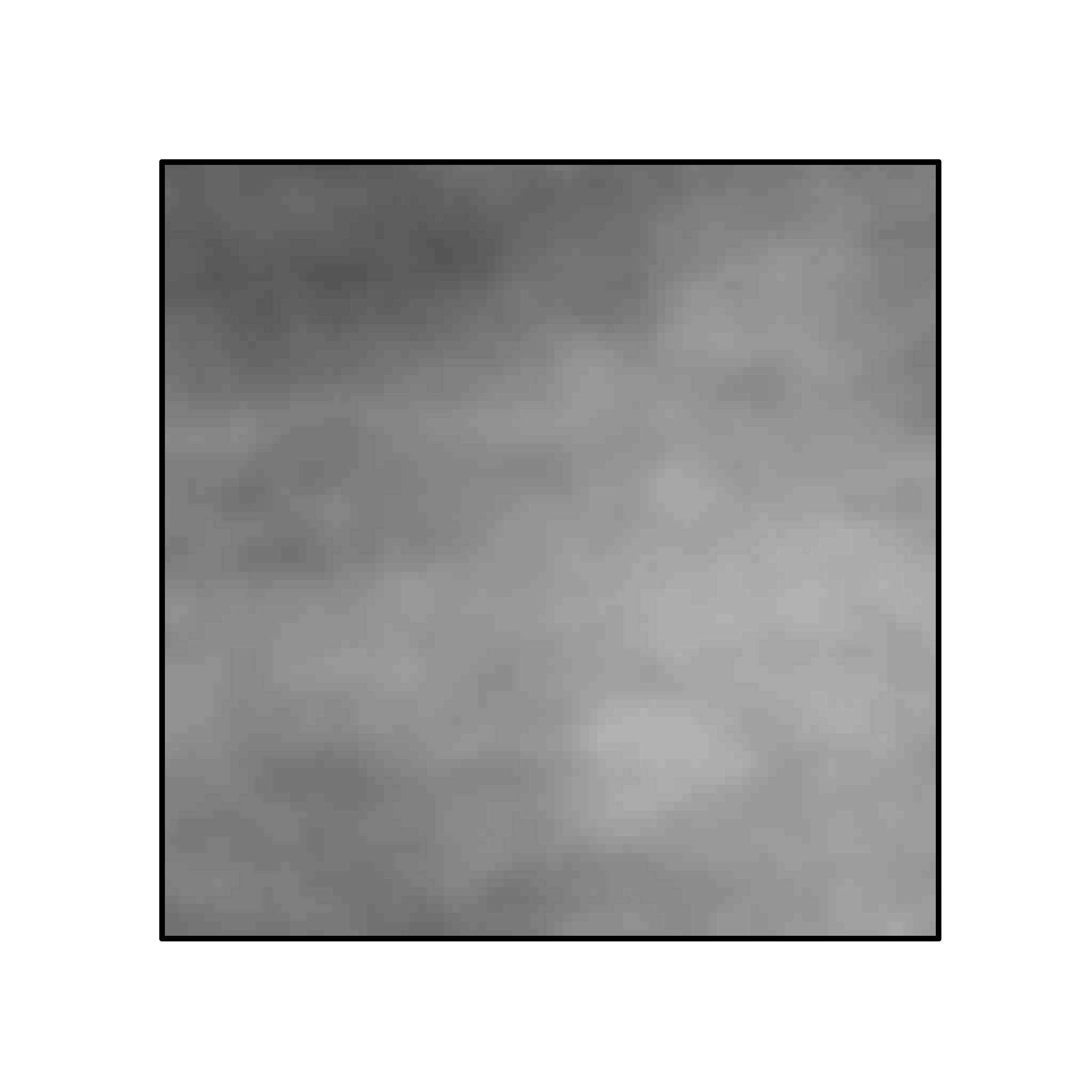}}} \quad
		\subfloat{\frame{\includegraphics[height=0.92\sizeFig]{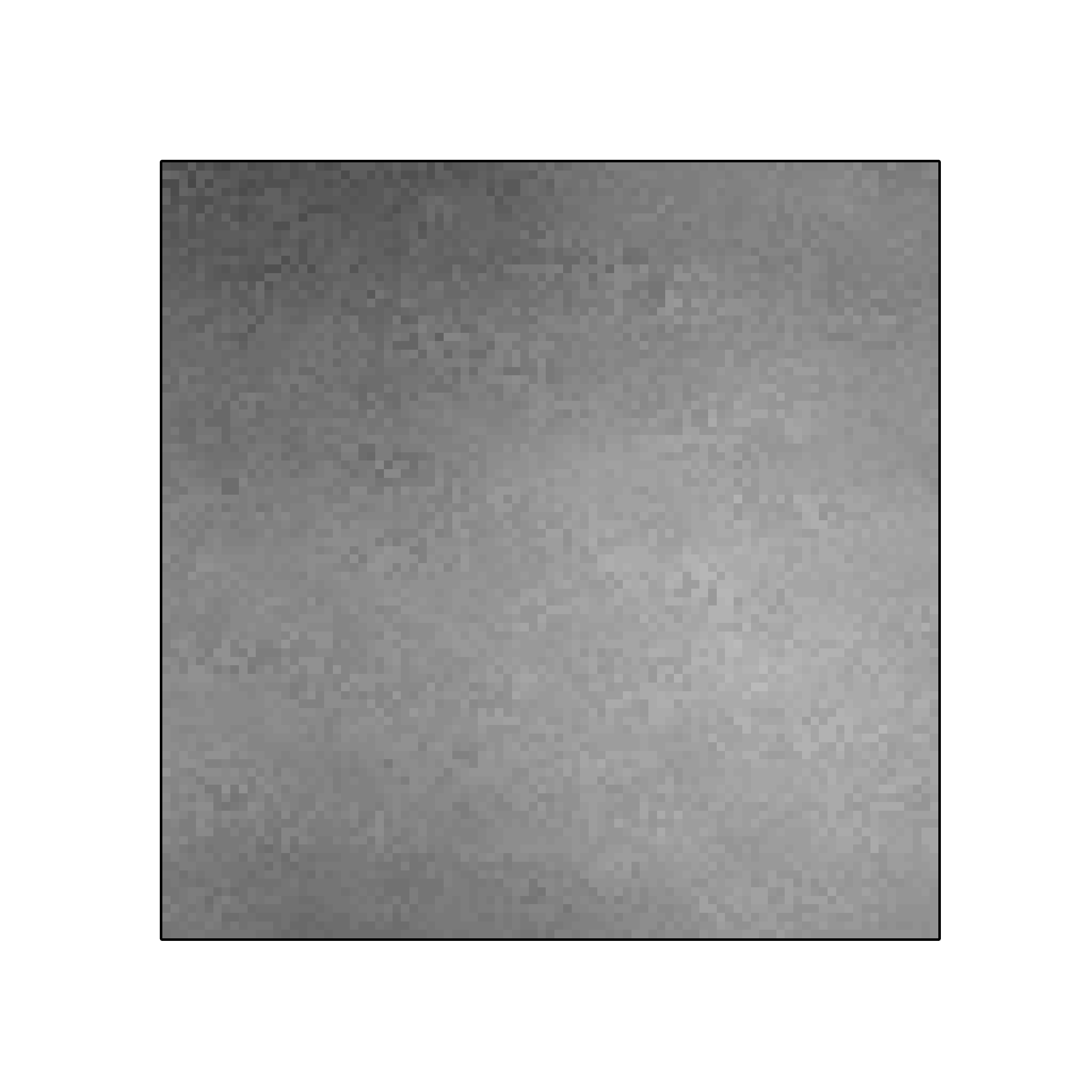}}} \quad
		\subfloat{\includegraphics[height=0.92\sizeFig]{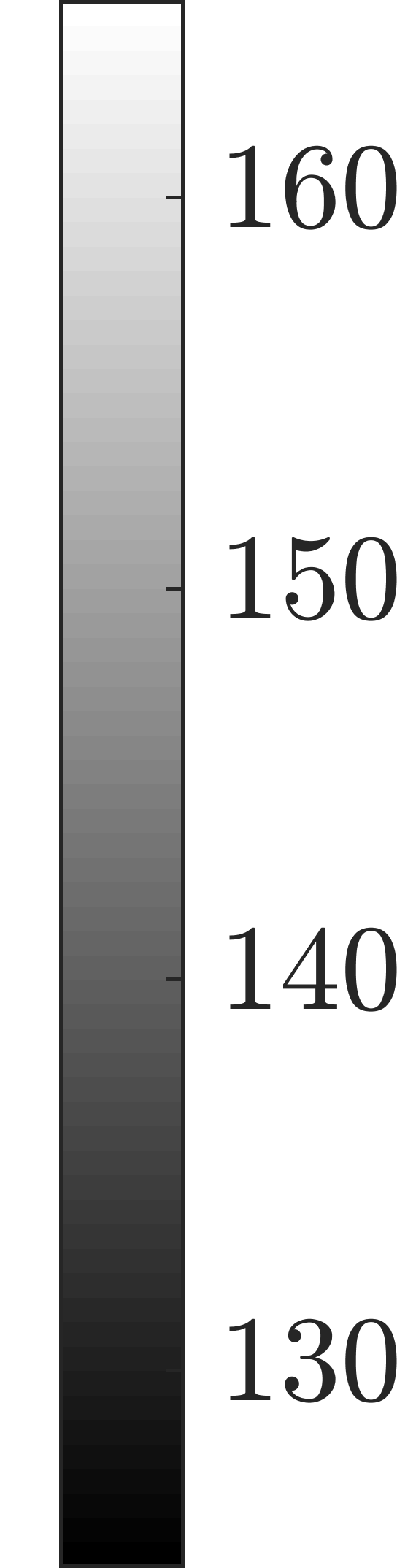}}\\ 
		\subfloat{\raisebox{0.4\sizeFig}{\rotatebox[origin=c]{90}{\vphantom{$\bs y$}Estimate $\bs{\tilde x}$}}} \quad
		\subfloat[\label{fig:results_real_phantom_a}Focused PSF]{\includegraphics[height=0.92\sizeFig]{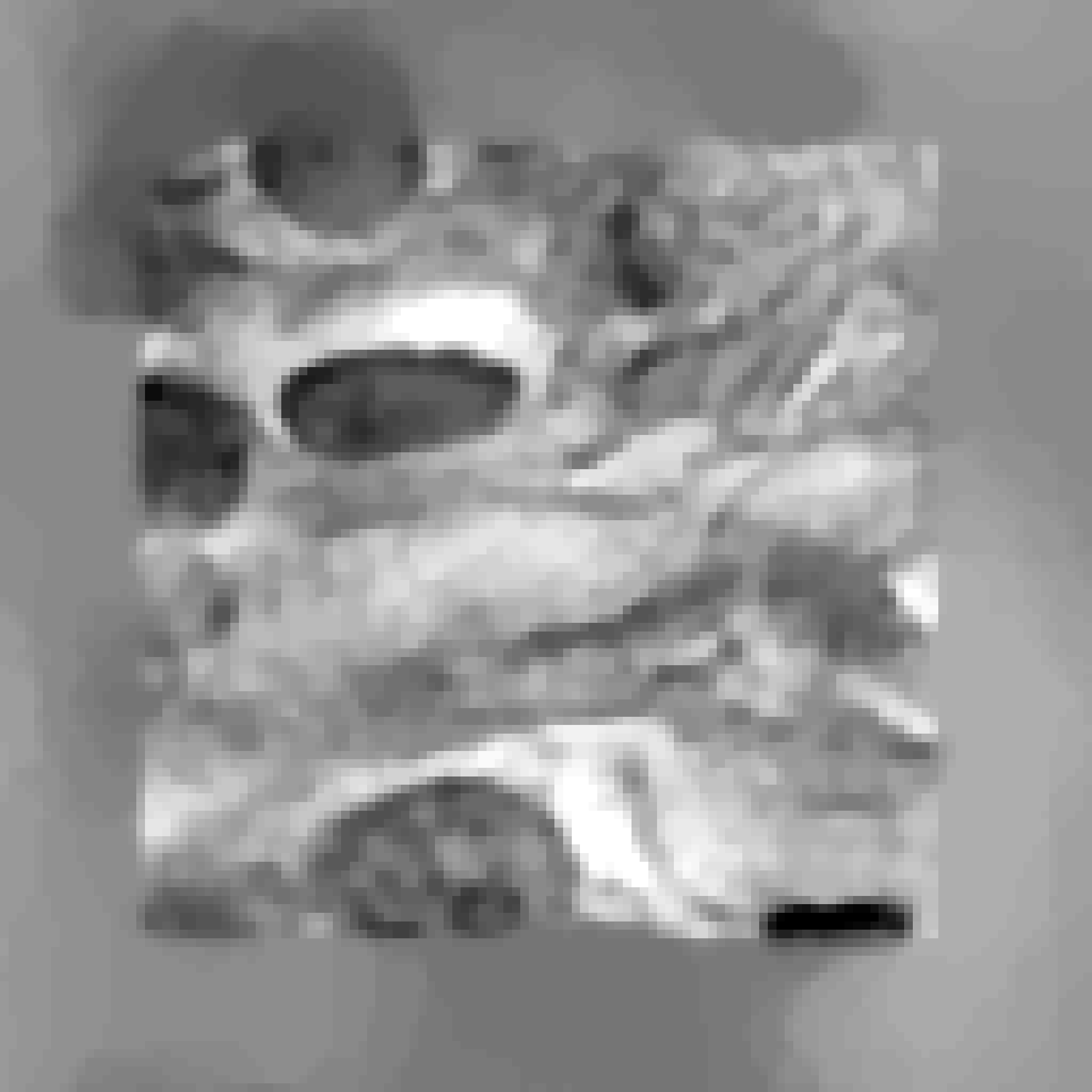}} \quad
		\subfloat[\label{fig:results_real_phantom_b}One random pattern]{\includegraphics[height=0.92\sizeFig]{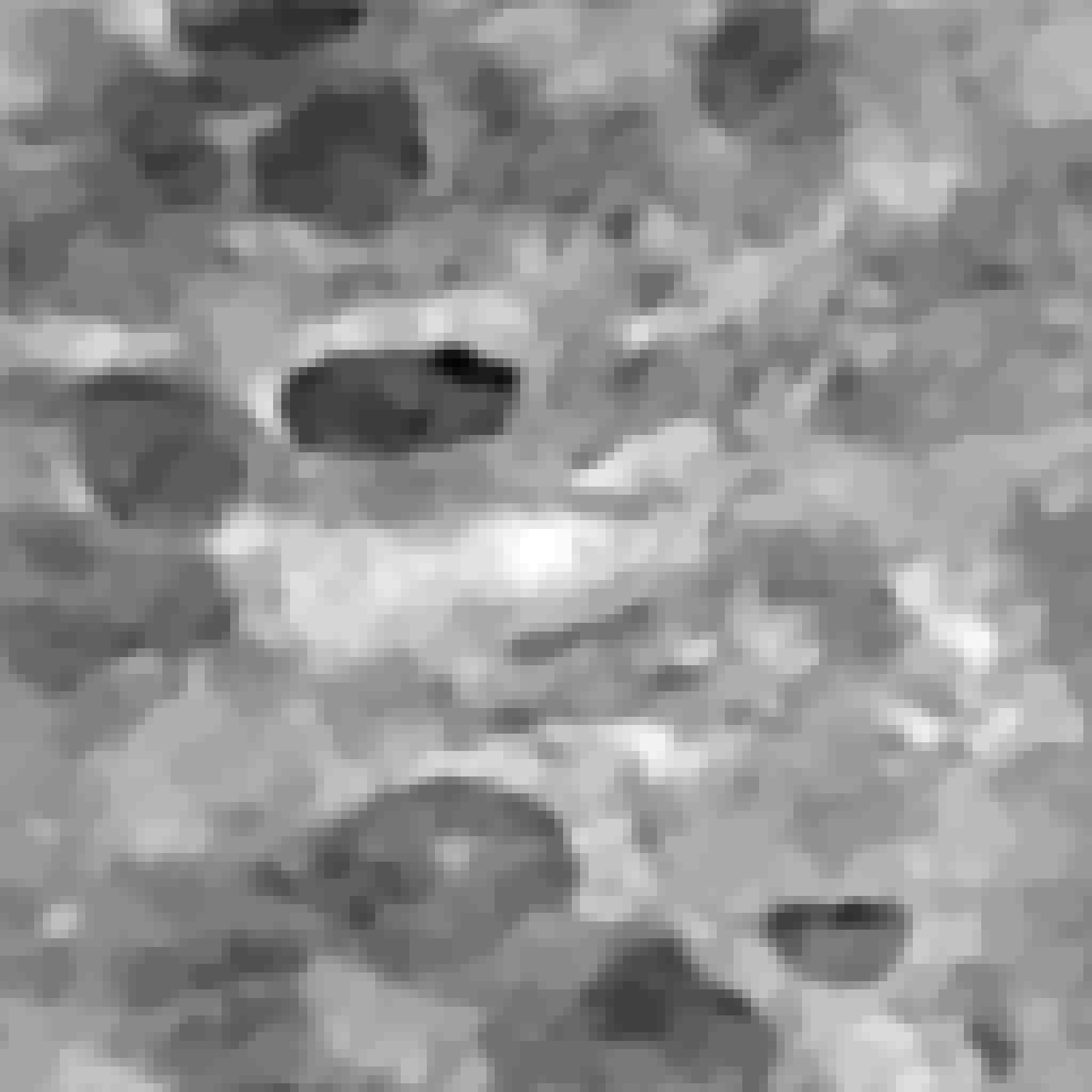}} \quad
		\subfloat[\label{fig:results_real_phantom_c}Four random patterns]{\includegraphics[height=0.92\sizeFig]{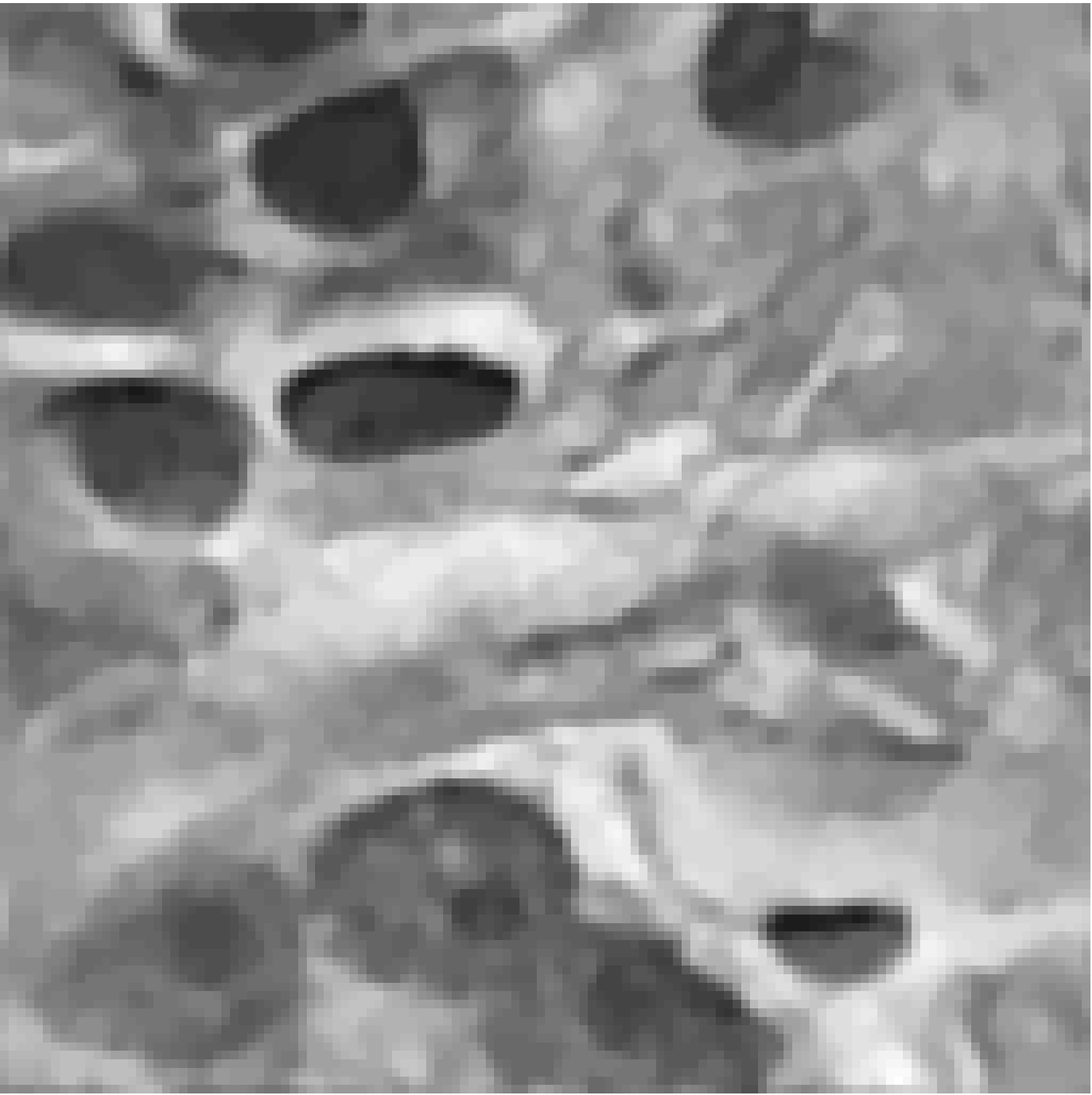}}\quad 
		\subfloat{\includegraphics[height=0.92\sizeFig]{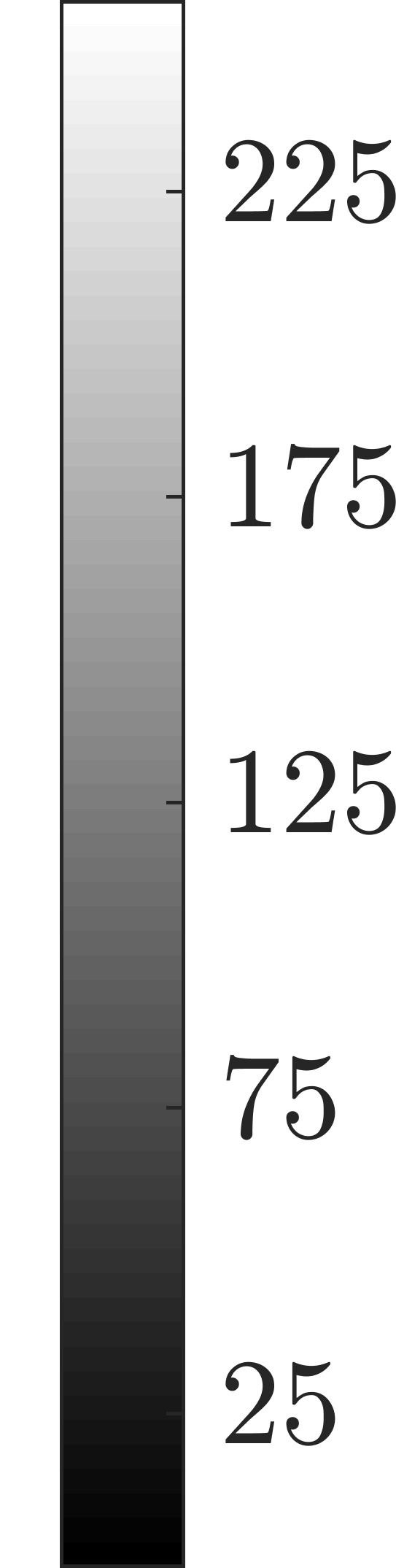}}
	\caption{Observations $\bs y$ (BSNR = 40 dB, $M/N=0.5$) (above) and estimate $\bs{\tilde x}$ (below). Observations obtained from raster scanning (left) and from illumination with random patterns (center and right).}
	\label{fig:results_images}
\end{figure}

Fig.\,\ref{fig:SNR-vs-compression_centered-samples} shows the SNR of reconstructed image $\tilde{\bs x}$ as a function of $M/N$ for focused PSF and three numbers of random patterns. When there is no compression, RS gives better results. However, when fewer measurements are available, using one random pattern already leads to better performance compared to RS. When we acquire more than $M=N/2$ observations (with $M<N$), using more patterns increases the SNR. Results for $M=N/2$ measurements are depicted in Fig.\,\ref{fig:results_images}.

\subsection{Conclusion}This preliminary work shows promising results for using CS approach in real acquisition setup. This technique does not require calibration and could either acquire the same FOV in less time or a bigger FOV in the same time. In future work, we will make deeper comparison with RS and we will investigate the statistics of the noise on real data and the choice of the image prior as well as the properties of the random patterns.


\begin{thebibliography}{10}

\bibitem{Milo2009}
R. Milo, P. Jorgensen, U. Moran, G. Weber, and M. Springer,
\newblock ``BioNumbers The database of key numbers in molecular and cell
  biology'',
\newblock Nucleic Acids Research, {\bf 38}(1):750--753, 2009.

\bibitem{Andresen2016}
E.~R. Andresen, S. Sivankutty, V. Tsvirkun, G. Bouwmans, and H. Rigneault,
\newblock ``Ultrathin endoscopes based on multicore fibers and adaptive optics:
  status and perspectives'',
\newblock Journal of Biomedical Optics, {\bf 21}(12):121506, 2016.

\bibitem{Andresen2013}
E.~R. Andresen, G. Bouwmans, S. Monneret, and H. Rigneault,
\newblock ``Two-photon lensless endoscope'',
\newblock Optics express, {\bf 21}(18):20713--20721, 2013.

\bibitem{Sivankutty2018}
S. Sivankutty, V. Tsvirkun, O. Vanvincq, G. Bouwmans, E. Andresen, and H. Rigneault,
\newblock ``Nonlinear imaging through a Fermat’s golden spiral multicore fiber'',
\newblock Optics letters, {\bf 43}(15):3638--3641, 2018.

\bibitem{Duarte2008}
M.~F. Duarte, M.~A. Davenport, D. Takhar, J.~N. Laska, T. Sun, K.~F. Kelly, and R.~G. Baraniuk,
\newblock ``Single-Pixel Imaging via Compressive Sampling'',
\newblock IEEE Signal Processing Magazine, {\bf 25}(2):83--91, 2008.

\bibitem{Candes2008}
E.~J. Cand\`es and M.~B. Wakin,
\newblock ``An Introduction To Compressive Sampling'',
\newblock IEEE Signal Processing Magazine, {\bf 25}(2):21--30, 2008.

\bibitem{Jacques2010a}
L. Jacques and P. Vandergheynst,
\newblock ``Compressed Sensing: ``When sparsity meets sampling'''',
\newblock Optical and Digital Image Processing, 2010.

\bibitem{Rudin1992}
L.~I. Rudin, S. Osher, and E. Fatemi,
\newblock ``Nonlinear total variation based noise removal algorithms'',
\newblock Physica D: Nonlinear Phenomena, {\bf 60}:259--268, 1992.

\bibitem{Chambolle2010}
A. Chambolle and T. Pock,
\newblock ``A First-Order Primal-Dual Algorithm for Convex Problems with
  Applications to Imaging'',
\newblock Journal of Mathematical Imaging and Vision, {\bf40}(1):120--145, 2010.

\bibitem{Kamilov2012}
U. Kamilov, E. Bostan, and M. Unser,
\newblock ``Wavelet shrinkage with consistent cycle spinning generalizes total
  variation denoising'',
\newblock IEEE Signal Processing Letters, {\bf 19}(4):187--190, 2012.

\bibitem{Almeida2013}
M.~S.~C. Almeida and M.~A.~T. Figueiredo,
\newblock ``Deconvolving images with unknown boundaries using the alternating
  direction method of multipliers'',
\newblock IEEE Transactions on Image Processing, {\bf22}(8):3074--3086, 2013.

\bibitem{Gonzalez2016a}
A. Gonz{\'{a}}lez, V. Delouille, and L. Jacques,
\newblock ``Non-parametric PSF estimation from celestial transit solar images
  using blind deconvolution'',
\newblock Journal of Space Weather and Space Climate, {\bf 6}:A1, 2016.

\bibitem{Almeida2010}
M.~S.~C. Almeida and L.~B. Almeida,
\newblock ``Blind and Semi-Blind Deblurring of Natural Images'',
\newblock IEEE Transaction on Image Processing, {\bf19}(1):36--52, 2010.

\bibitem{Bredies2010}
K. Bredies, K. Kunisch, and T. Pock,
\newblock ``Total Generalized Variation'',
\newblock SIAM Journal on Imaging Sciences, {\bf3}(3):492--526, 2010.

\bibitem{Parikh2013}
N. Parikh and S. Boyd,
\newblock ``Proximal algorithms'',
\newblock Foundations and Trends in optimization, {\bf 1}(3):123--231, 2013.

\bibitem{Almeida2013a}
M.~S.~C. Almeida and M.~A.~T. Figueiredo,
\newblock ``Parameter estimation for blind and non-blind deblurring using
  residual whiteness measures'',
\newblock IEEE transactions on image processing : a publication of the
  IEEE Signal Processing Society, {\bf 22}(7):2751--63, 2013.
  
\bibitem{Matlab}
MATLAB Release 2016a, The MathWorks, Inc., Natick, Massachusetts, United States.

\end{thebibliography}
\end{document}